**A multimodal multiplex of the mental lexicon for multilingual individuals**

Maria Huynh and Wilder C. Rodrigues

Faculty of Humanities

Utrecht University

Artificial Intelligence

Ana Bosnic, Rick Nouwen, Shalom Zuckerman

November 7, 2025

# 1 Introduction

Historically, bilingualism was often perceived as an additional cognitive load that could hinder linguistic and intellectual development. However, over the last three decades, this view has changed considerably. Numerous studies have aimed to model and understand the architecture of the bilingual word recognition system Dijkstra and van Heuven (2002), investigating how parallel activation operates in the brain and how one language influences another Kroll et al. (2015). Increasingly, evidence suggests that multilinguals, individuals who speak three or more languages, can perform better than monolinguals in various linguistic and cognitive tasks, such as learning an additional language Abu-Rabia and Sanitsky (2010).

This research proposal focuses on the study of the mental lexicon and how it may be structured in individuals who speak multiple languages. Building on the work of Stella et al. (2018), who investigated explosive learning in humans using a multiplex model of the mental lexicon, and the Bilingual Interactive Activation (BIA+) framework proposed by Dijkstra and van Heuven (2002), the present study applies the same multilayer network principles introduced by Kivelä et al. (2014).

Our experimental design extends previous research by incorporating multimodality into the multiplex model, introducing an additional layer that connects visual inputs to their corresponding lexical representations across the multilingual layers of the mental lexicon.

In this research, we aim to explore how a heritage language influences the acquisition of another language. Specifically, we ask: **Does the presence of visual input in a translation task influence participants' proficiency and accuracy compared to text-only conditions?**

# 2 Theoretical Background

## 2.1 The multiplex model of the mental lexicon

Building on the multilayer network framework introduced by Kivelä et al. (2014), Stella et al. (2018) proposed the *multiplex model of the mental lexicon*, in which linguistic relationships such as semantic, taxonomic, and phonological connections are represented as distinct yet interconnected layers. These layers interact dynamically to form a complex, integrated structure capable of supporting language learning and retrieval. The model conceptualises language learning as an emergent property of this interconnectivity, giving rise to what the authors describe as an *explosive learning phenomenon*: a stage in which word connectivity and navigability increase rapidly, reflecting a more integrated and efficient language system.

Within this framework, the emergence of the *Largest Viable Cluster* (LVC) is a central mechanism. The LVC represents a subset of words that are highly interconnected across multiple linguistic layers and are, consequently, more easily recognised, memorised, and learned. Words within the LVC tend to be used more frequently and acquired earlier, indicating that the lexicon's structural organisation plays a critical role in language development.

Two linguistic properties, *polysemy* and *concreteness*, further influence how the LVC evolves. Polysemy refers to a word having multiple meanings (e.g., "head" in *head of a person* or *head of a department*), while concreteness describes how tangible or imageable a word is. Empirical research shows that children tend to learn more concrete words earlier in development. To investigate the contribution of these factors, Stella et al. (2018) applied a partial reshuffling null model, in which the network's structural topology (the pattern of word-to-word connections) is preserved, but lexical properties such as frequency, concreteness, and polysemy are randomised. This controlled randomisation breaks natural correlations in the real data, allowing an assessment of how these features independently affect the formation of the LVC.

When comparing the real language network to the reshuffled model, the authors observed that polysemy had a substantially greater effect on the emergence of the LVC than concreteness—the gap in polysemy scores was approximately five times larger. They also found that around the age of eight, the LVC exhibited a peak in highly polysemous words compared to the LVC at age twenty, a pattern that was absent in the reshuffled model. These findings suggest that polysemy, despite introducing semantic ambiguity, is a crucial feature that enhances the connectivity of the mental lexicon and facilitates the rapid expansion of linguistic knowledge. Moreover, polysemy appears to be a universal property across languages, reflecting its fundamental role in lexical organisation.

Together, these insights reinforce the notion that language learning and lexical access rely on interconnected networks rather than isolated word representations. The multiplex model therefore provides a theoretical foundation for understanding how multimodal and multilingual input might further enhance network connectivity, an assumption that underlies the present study's hypothesis.

## 2.2 Multilayer networks

In traditional network theory, systems are typically described using a single layer of connections. However, complex systems across the social, biological, physical, information, and engineering sciences often involve multiple, interdependent types of relationships. Kivelä et al. (2014) formalised this idea by introducing the concept of *multilayer networks*, aiming to address the explosion of terminology resulting from numerous studies that used inconsistent definitions. They described the lack of consensus as "extremely problematic," noting that different terms were often used for the same mathematical object, while the same term was applied to different objects. The primary goal of their research was to resolve this fragmentation by proposing a unified framework capable of integrating these conflicting concepts.

Their framework defined a multilayer network $M$ as a quadruplet $M = (V_M, E_M, V, L)$, where:

- $V$ is the set of all unique nodes (entities);

- $L$ is the sequence of sets of elementary layers for all aspects;

- $V_M$ is the set of existing node–layer tuples (allowing nodes to be absent from some layers);

- $E_M$ is the set of edges connecting pairs of node–layer tuples.

The quadruplet definition of a multilayer network is illustrated in Figure 1, showing how two-layered networks can be interconnected through different layers ($L$) and edges ($E_M$). It is important to note that the edges connecting pairs of node–layer tuples enable navigation across the entire newly formed multilayer network. This extensive connectivity underlies the multiplex model of the mental lexicon proposed by Stella et al. (2018), which employs the multilayer framework to support and visualise their hypothesis regarding the phenomenon of explosive learning in humans.

In psycholinguistics, this structure aligns well with the notion that lexical representations are distributed across semantic, phonological, and orthographic aspects. When applied to a multimodal concept with both visual and textual input, multilayer networks can capture how activation in one modality may spread to another. This makes the framework particularly suitable for modelling the interconnected nature of multimodal lexical access.

This theoretical foundation therefore provides the structural basis for the multiplex lexical model proposed by Stella et al. (2018), which applies the multilayer approach specifically to the mental lexicon.

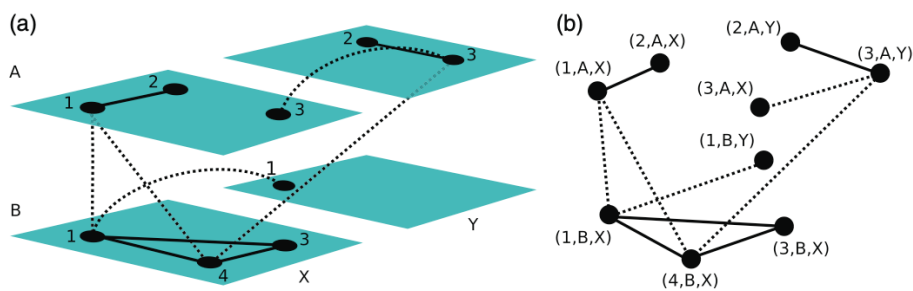

**Figure 1**
*Source: Kivelä et al. (2014). Schematic representation of a multilayer network. (a) Each layer represents a different type of relationship (e.g. friendship, family, etc.), and nodes can appear in one or more layers. (b) The multilayer structure can be unfolded into a single interconnected representation, highlighting the interdependence between layers.*

## 2.3 The architecture of the bilingual word recognition system

Over the past decades, research on bilingualism, and particularly on the bilingual mental lexicon, has expanded rapidly. This growth reflects the unique challenges bilinguals face that monolinguals do not, such as distinguishing between words from two different languages during reading or listening. To better understand these challenges, it is essential to examine how words from multiple languages are stored, accessed, and selected during comprehension. Early models of bilingual word recognition (Dijkstra and van Heuven (1998)) proposed that the two languages of bilinguals were either separately organised, with independent lexicons, or selectively accessed, with only the target language being activated during recognition. However, subsequent research provided strong evidence that both languages remain simultaneously active during word recognition, even when only one language is in use (Dijkstra and van Heuven (2002)).

To account for this phenomenon, Dijkstra and van Heuven (2002) introduced the *Bilingual Interactive Activation Plus* (BIA+) model, an extension of their earlier *Bilingual Interactive Activation* (BIA) framework. The BIA+ model is a computational model designed to explain how bilingual individuals recognise written words through orthographic representations. It incorporates and extends the original BIA architecture to account for a wider range of empirical findings on bilingual word recognition. The model assumes that bilinguals share a single, integrated mental lexicon in which orthographic, phonological, and semantic representations from both languages interact dynamically. When a bilingual encounters a word, lexical candidates from both languages are automatically activated, producing cross-linguistic competition before the correct word is identified—a process known as *language non-selective lexical access.*

The BIA+ model distinguishes between two interconnected subsystems with distinct roles. The *word identification system* automatically recognises and activates lexical representations from both languages, even when only one language is being used. For example, the English word *room* activates its English meaning ("a space in a house") but also co-activates the Dutch word *room*, meaning "cream." The *task/decision system*, in contrast, operates at a higher cognitive level, helping individuals select the appropriate language for response based on contextual cues or task demands. Earlier theories suggested that task instructions could directly influence word recognition itself; however, the BIA+ model clarifies this misconception by distinguishing automatic lexical activation from conscious decision-making.

Another important refinement in the BIA+ model concerns the concept of *language nodes*. In the original BIA model, language nodes functioned as control switches—turning one language on and the other off. In the BIA+ model, however, they act more like *labels*, marking which language a word belongs to without directly controlling activation. This revision brings the model closer to real bilingual behaviour: even when bilinguals are instructed to use only one language, cross-language effects such as cognate facilitation and false-friend interference persist. If the brain could fully suppress the non-target language, such effects would not occur.

Overall, the BIA+ model provides strong evidence for the interactive and non-selective nature of bilingual lexical access. Its principles are directly relevant to the present study, as the same mechanisms of parallel activation and cross-language competition are expected to operate in multilingual individuals exposed to multimodal stimuli. The model therefore offers a theoretical foundation for understanding how visual and linguistic inputs might jointly influence lexical retrieval and translation performance.

## 2.4 Bilingualism, Mind, and Brain

The dynamic interplay that arises from the constant interaction between languages within a single cognitive system has direct consequences for the mind, the brain, and language processing itself. This interaction provides a unique lens through which researchers can observe fundamental mechanisms of language and cognition that often remain hidden in monolingual speakers.

The long-standing but inaccurate notion, famously summarised by François Grosjean's statement that a bilingual is "two monolinguals in one person," does not align with contemporary evidence. Kroll et al. (2015) challenges this view through what they define as the three core discoveries in bilingual research, emphasising that bilinguals integrate rather than separate their linguistic systems.

From the three core discoveries outlined by Kroll et al. (2015): (1) Parallel Activation: The Constant Coexistence of Languages; (2) Bidirectional Influence: A Dynamic and Permeable Language System; and (3) Cognitive and Neural Consequences: The Bilingual Advantage. The first, *Parallel Activation*, is the most directly relevant to the present research. This core concept emphasises that multiple languages are simultaneously active in the bilingual mind, leading to continuous interaction between linguistic systems.

Evidence for parallel activation comes from a diverse range of experimental tasks and methodologies. One important approach involves the processing of language-ambiguous cognates and homographs. Cognates are words that share similar form and meaning across languages (e.g., *piano* in English and Spanish) and have been shown to facilitate performance in both word recognition and word production tasks (e.g., Costa et al. (2000), Dijkstra (2005)). Homographs, on the other hand, share lexical form but differ in meaning across languages (e.g., Spanish *carpeta*, meaning 'folder,' not 'carpet'), and tend to induce cross-language interference during processing or production.

In the context of the proposed multimodal multiplex model of the mental lexicon, we expect parallel activation to play a central role. However, the addition of a visual modality (i.e., schematic or pictorial input) is predicted to reduce the interference typically observed with homographs. Instead, the visual cues are expected to facilitate translation for multilingual participants by strengthening the mapping between conceptual and linguistic representations across languages.

## 2.5   Advantages of Bilinguals Over Monolinguals in Learning a Third Language

The study conducted by Abu-Rabia and Sanitsky (2010) investigated the contribution of bilingualism to trilingualism by examining how mastery of two different writing systems (orthographies) could influence the learning of a third language. The researchers focused on the transfer of cognitive and linguistic skills from a student's first language (L1) and second language (L2) to a third, foreign language (L3/FL). Their sample consisted of two groups of sixth-grade students from Israeli schools who were studying English as a foreign language (either as their second or third language). The first group included Russian–Israeli children whose native language was Russian and whose second language was Hebrew, while the control group consisted of native Hebrew speakers.

The study highlighted several structural discrepancies between the languages involved. According to Abu-Rabia and Sanitsky (2010), these include major variations in writing systems, morphology, phonetics, gender marking, syllable structure, and alphabet size. Despite these challenges, the Russian–Israeli group performed better on most English tasks compared with the native Hebrew-speaking group. Multivariate analyses of variance (MANOVA) revealed significant differences between the groups, with the trilingual native Russian speakers outperforming the Hebrew speakers on multiple English language measures.

These results indicate a strong transfer of cognitive and linguistic skills across languages. This finding supports the present study by demonstrating how multilingual experi-ence facilitates the acquisition of a new language, reinforcing the assumption that additional input (i.e., visual stimuli) may further strengthen language learning and processing.

## 2.6 The Wernicke area: Modern evidence and a reinterpretation

Recent studies (Binder (2015)) provide biological grounding for the proposed multimodal extension of the lexical network. As shown in Figures 2a–2b, the supramarginal gyrus (SMG) and the inferior parietal lobe, which include Wernicke's area and Brodmann area 39, respectively, are critically involved in phonological retrieval, semantic processing, and the comprehension of language. These regions are anatomically adjacent to the occipitotemporal cortex, which contains the visual letter perception system, thereby forming a close interface between visual perception and language comprehension.

This spatial and functional proximity suggests that visual input can directly influence phonological and semantic activation, as suggested by Binder (2015). In this sense, the multimodal pathways linking occipitotemporal visual regions to parietal language areas may support the same kind of cross-layer connectivity proposed in our multimodal multiplex for the mental lexicon. Thus, visual stimuli are expected to facilitate lexical retrieval and strengthen cross-linguistic activation, providing neurocognitive support for the hypothesis that multimodal input enhances language processing in multilingual individuals.

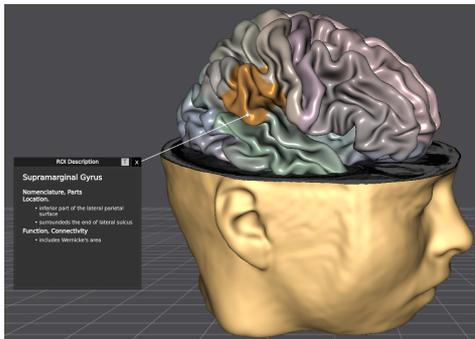

(a) *The supramarginal gyrus, located in the inferior parietal region, surrounds the end of the lateral sulcus and includes Wernicke's area. It plays a central role in phonological retrieval and integration with auditory and visual input. Source: BrainTutor.*

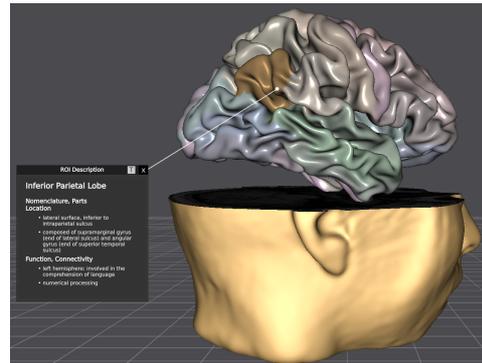

(b) *The inferior parietal lobe, composed of the supramarginal and angular gyri (Brodmann area 39), is involved in language comprehension and semantic processing, and lies adjacent to the occipitotemporal visual regions. Source: BrainTutor.*

**Figure 2**
*Neurocognitive regions supporting multimodal lexical activation. (a) The supramarginal gyrus and Wernicke's area are involved in phonological retrieval; (b) the inferior parietal lobe and Brodmann area 39 are involved in semantic comprehension.*

# 3 Research Question & Hypothesis

## 3.1 Research Question

**Does the presence of visual input in a translation task influence participants' proficiency and accuracy compared to text-only conditions?**

This question aims to clarify the role of multimodal input in language processing and learning. Understanding how visual and textual information interact during translation may provide insight into the mechanisms by which multimodal stimuli enhance lexical access, memory consolidation, and the acquisition and production of additional languages.

## 3.2 Hypothesis

Building on the multiplex lexical model proposed by Stella et al. (2018), this study extends the representation of the mental lexicon by incorporating an additional visual layer that grounds linguistic nodes in perceptual experience (see Figure 3). This multimodal extension assumes that language processing involves interconnected lexical networks across modalities, where visual representations can amplify the influence of both the native and heritage languages on cognitive processing during third- or fourth-language production.

It is therefore hypothesised that visual input will enhance translation proficiency and accuracy across participant groups. More specifically, the following outcomes are expected:

- **H1:** Multilingual participants in the Visual–Textual (VT) condition are expected to achieve higher translation proficiency and accuracy than multilinguals in the Only–Text (OT) condition.

- **H2:** Bilingual participants in the VT condition are expected to outperform bilinguals in the OT condition.

- **H3:** Multilinguals in the VT condition are predicted to perform better overall than bilinguals in the same condition, reflecting the combined influence of multiple language systems and multimodal input.

This expectation is supported by evidence from Kroll et al. (2015), who demonstrated that not only does the first language (L1) influence the second language (L2), but the L2 also begins to affect the L1 once bilinguals reach sufficient proficiency.

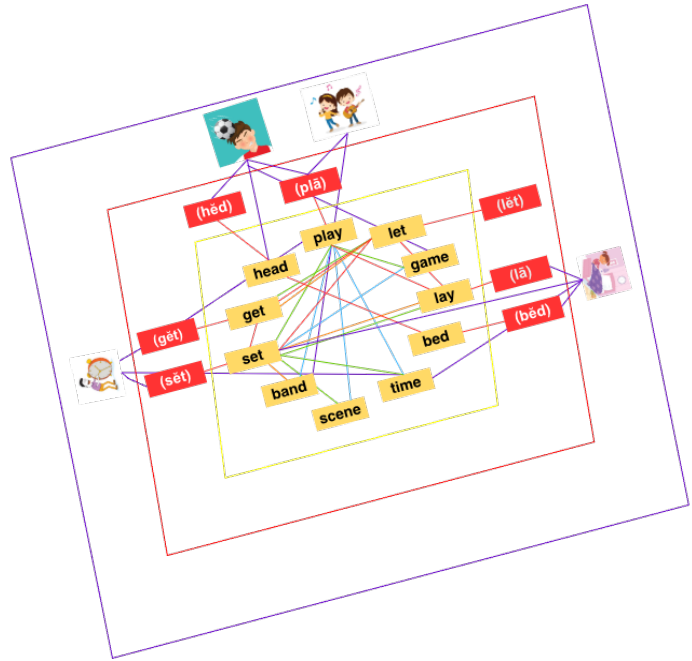

**Figure 3**

*Proposed multimodal multiplex model of the mental lexicon, here being compared to the original model by Stella et al., 2018 illustrated in Figure 4. The addition of a visual layer introduces perceptual grounding and cross-modal connections that enhance lexical activation.*

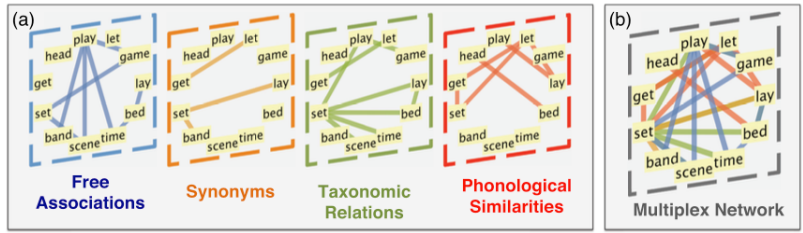

**Figure 4**

*Source: Stella et al. (2018). (a) Visual representation of a subset of the multiplex lexical representation (MLR) for adults with $N = 8531$ words and four types of word relationships forming individual layers: free associations, synonyms, taxonomic relations, and phonological similarities. (b) Multiplex visualisation as an edge-coloured network.*

# 4 Method

## 4.1 Participants

The experiment will include participants aged between 12 and 16 years. This age range is chosen to ensure that all participants are alphabetised in their native language (L1) and have received sufficient exposure to their second language (L2). Setting the upper limit at 16 years also prevents the inclusion of participants who may already be enrolled in a university degree programme. This minimises potential disparities in English proficiency and helps maintain a more homogeneous sample in terms of language experience.

Participants will be divided into two main groups based on linguistic background:

- **Bilingual (L2) group:** native speakers of Dutch (L1) and proficient in English (L2);

- **Multilingual (L*) group:** native speakers of Dutch (L1), proficient in English (L2), and with some fluency in Portuguese (heritage language, HL).

Each main group will be further subdivided according to the type of stimuli presented:

- **VT (Visual–Textual)**: participants receive multimodal stimuli (image and text);

- **OT (Only–Text)**: participants receive text-only stimuli.

This results in four subgroups (L2–VT, L2–OT, L*–VT, L*–OT). Both VT subgroups will serve as control groups, as the experiment aims to assess whether multimodal input leads to enhanced translation accuracy and proficiency compared with text-only conditions.

To minimise potential confounding factors, participants' results will also be aggregated by age brackets within the 12–16 range.

## 4.2 Materials

The experimental materials consist of two sets of linguistic stimuli and a corresponding set of visual stimuli for the multimodal condition.

### 4.2.1 Linguistic stimuli.

Fifteen Dutch sentences are selected to represent everyday language use, including both simple and complex syntactic structures. Each sentence is accompanied by its grammatically correct translations in English, which will later be used as reference sentences for the embedding-based proficiency analysis. The stimuli are the same for all participants to allow cross-group comparisons.

### 4.2.2 Visual stimuli.

For the Visual–Textual (VT) condition, each sentence is paired with a corresponding image depicting the main concept or action described by the sentence (e.g., a child throwing a ball, a dog running). The images are selected to be culturally neutral, age-appropriate, and easily interpretable, ensuring that visual information supports comprehension rather than introducing ambiguity.

### 4.2.3 Recording and transcription tools.

Participants' spoken translations are recorded using standardised audio equipment to ensure consistent sound quality across sessions. The recordings are automatically transcribed using speech-to-text software and manually reviewed for minor spelling corrections while preserving syntactic and semantic accuracy.

### 4.2.4 Embedding and analysis tools.

Both reference translations and participant transcripts are converted into sentence embeddings using a pre-trained multilingual transformer model (Stankevicius and Lukosevicius (2025)). The similarity between reference and participant embeddings is calculated using cosine distance to quantify translation accuracy and overall proficiency.

## 4.3 Procedure

As shown in Figure 5, each experimental trial follows a fixed sequence. The experiment begins with a fixation cross presented at the centre of the screen. Although no eye-tracking is involved, the fixation cross serves to direct participants' attention and ensure that they are focused on the screen before the stimulus appears. Participants first view the fixation cross for 200 ms, followed by the stimulus presentation. In the Visual–Textual (VT) condition, an image is displayed for 1 second and is followed by the corresponding Dutch sentence for 2 seconds. In the Only–Text (OT) condition, only the sentence is shown for the same duration.

After the presentation phase, a red recording cue appears on screen for 4 seconds, during which participants verbally translate the Dutch sentence into English. Each session lasts approximately 15 minutes and consists of 15 trials. Before starting, all participants complete a short practice round to familiarise themselves with the procedure.

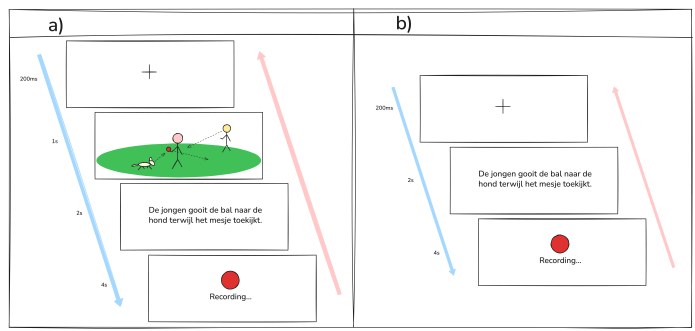

**Figure 5**
*Stimulus presentation timeline for the two experimental conditions. (a) Visual–Textual (VT) condition, where an image and the corresponding Dutch sentence are shown sequentially before the recording prompt. (b) Only–Text (OT) condition, where participants view only the sentence before recording. Timings (in milliseconds and seconds) indicate stimulus duration and response window.*

# 5 Expected Results

It is expected that participants in the Visual–Textual (VT) condition will show higher translation proficiency and accuracy than those in the Only–Text (OT) condition, in line with the hypotheses. Multilingual participants are anticipated to obtain the highest similarity scores between their translations and the reference embeddings, particularly in the VT condition, whereas bilingual participants are expected to score lower overall.

These outcomes would manifest as a main effect of modality (VT > OT) and an interaction between modality and language group, with the largest gains for multilinguals under multimodal presentation. Cosine-similarity measures are predicted to reflect these effects, revealing smaller semantic distances for VT trials.

Such a pattern would confirm that visual input facilitates cross-linguistic activation and reduces interference, providing behavioural evidence for the proposed multimodal multiplex model of the mental lexicon.

## 5.1 Analysis

Recordings of participants' spoken responses will be transcribed and pre-processed. Minor spelling errors will be corrected to ensure consistent lexical forms while preserving syntactic and semantic accuracy.

A set of reference translations will be created by highly proficient bilingual speakers (Dutch L1, English L2). These reference sentences will be converted into multilingual sentence embeddings, which serve as target representations for comparison.

Participant transcripts will also be converted into embeddings. Translation proficiency will then be quantified by computing the cosine similarity between participant and reference embeddings. Higher similarity scores will indicate greater semantic accuracy and fluency in translation performance.

# 6 Discussion and Conclusion

If the expected results are confirmed, they would provide empirical support for the proposed multimodal multiplex model of the mental lexicon. Enhanced performance in the Visual–Textual (VT) condition would suggest that visual cues strengthen cross-linguistic activation and reduce interference, leading to more efficient lexical retrieval.

Such findings would align with the principles of connectionism and small-world network theory, supporting the view that the mental lexicon operates as a highly interconnected, multimodal network rather than a modular or language-specific system. They would also reinforce Kroll et al. (2015) framework on parallel activation, showing that multilingual experience extends beyond linguistic domains into perceptual processing.

In broader terms, these results would contribute to understanding how multimodal input facilitates language learning and production, with potential implications for educational settings and second-language teaching methodologies. The study would thus bridge computational and psycholinguistic perspectives, highlighting the value of network-based models in explaining human language acquisition and processing.

# 7 Potential Problems

Several potential challenges may affect the present study. First, the age range of participants (12–16 years) may introduce variability in reading speed, working memory, and overall language proficiency, as these abilities continue to develop throughout adolescence.

Second, individual differences in English proficiency and prior exposure to the language, through schooling, media, or personal use, could influence translation performance across groups.

Third, the linguistic material comprises a relatively small sample of fifteen sentences, which may limit the generalisability of the results and may not fully capture participants' overall translation abilities.

Future research could address these limitations by narrowing the participant age range, administering a standardised language proficiency assessment, and increasing the number of stimuli to ensure broader representativeness.

# 8 Future Research

Future studies could replicate this experiment using language pairs that differ substantially in linguistic structure, such as syntax, morphology, or orthography. This would help determine whether the effects of multimodal input vary across typologically distinct languages.

Additionally, future research could employ neuroimaging methods such as EEG or fMRI to investigate neural activation patterns in multilinguals during multimodal translation tasks. Such approaches would enable researchers to examine whether visual stimuli strengthen functional connectivity between visual and language processing regions, including the occipitotemporal cortex and Wernicke's area.

# References

Abu-Rabia, S., & Sanitsky, E. (2010). Advantages of bilinguals over monolinguals in learning a third language. *Bilingual Research Journal*, *33*, 173–199. https://doi.org/10.1080/15235882.2010.502797

Binder, J. R. (2015). The wernicke area: Modern evidence and a reinterpretation. *Neorology Journals*, *84*, 2170–2175. https://doi.org/10.1212/WNL.0000000000002219

Costa, A., Caramazza, A., & Galles, N. S. (2000). The cognate facilitation effect: Implications for models of lexical access. *Journal of Experimental Psychology. Learning, Memory and Cognition*, *26(5)*, 1283–1296. https://doi.org/10.1037/0278-7393.26.5.1283

Dijkstra, T. (2005). Bilingual visual word recognition and lexical access. *In Judith F. Kroll & Annette M. B. de Groot (Eds.), Handbook of bilingualism: Psycholinguistic approaches*, 179–201. https://doi.org/10.1037/0278-7393.26.5.1283

Dijkstra, T., & van Heuven, W. J. B. (1998). The bia-model and bilingual word recognition. *In J. Grainger & A. M. Jacobs (Eds.), Localist connectionist approaches to human cognition*, 189–225.

Dijkstra, T., & van Heuven, W. J. B. (2002). The architecture of the bilingual word recognition system: From identi®cation to decision. *Bilingualism: Language and Cognition*, *5*, 175–197. https://doi.org/10.1017/S1366728902003012

Kivelä, M., Arenas, A., Barthelemy, M., Gleeson, J. P., Moreno, Y., & Porter., M. A. (2014). Multilayer networks. *Journal of Complex Networks*, *2*, 203–271.

Kroll, J., Dussias, P. E., Bice, K., & Perrotti, L. (2015). Bilingualism, mind, and brain. *Annual Review of Linguistics*, 377–394. https://doi.org/10.1146/annurev-linguist-030514-124937

Stankevicius, L., & Lukosevicius, M. (2025). Extracting sentence embeddings from pretrained transformer models. *ArXiv - https://arxiv.org/pdf/2408.08073v2.*

Stella, M., Beckage, N. M., Brede1, M., & Domenico, M. D. (2018). Multiplex model of mental lexicon reveals explosive learning in humans. *Scientific Reports*. https://doi.org/10.1038/s41598-018-20730-5